# Reinforcement Learning-Guided NSGA-II Enhanced with Gray Relational Coefficient for Multi-Objective Optimization: Application to NASDAQ Portfolio Optimization

Zhiyuan Wang [a, *], Qinxu Ding [a], Ding Ding [a, *] Siying Zhu [a], Jing Ren [a], Yue Wang [a], Chong Hui Tan [a]

[a] School of Business, Singapore University of Social Sciences, Singapore 599494, Singapore

* Corresponding authors: Zhiyuan Wang, Email: zywang@suss.edu.sg; Ding Ding, Email: dingding@suss.edu.sg

**Abstract**

In modern financial markets, decision-makers increasingly rely on quantitative methods to navigate complex trade-offs among multiple, often conflicting objectives. This paper addresses constrained multi-objective optimization (MOO) with an application to portfolio optimization for minimizing risk and maximizing return. To this end, and to address existing gaps, we propose a novel reinforcement learning (RL)-guided non-dominated sorting genetic algorithm II (NSGA-II) enhanced with gray relational coefficients (GRC), termed RL-NSGA-II-GRC, which combines an RL agent controller and GRC-based selection to improve the convergence and diversity of the Pareto-optimal fronts. The agent adapts key evolutionary parameters online using population-level metrics of hypervolume, feasibility, and diversity, while the GRC-enhanced tournament operator ranks parents via a unified score simultaneously considering dominance rank, crowding distance, and geometric proximity to ideal reference. We evaluate the framework on the Kursawe and CONSTR benchmark problems and on a NASDAQ portfolio optimization application. On the benchmarks, RL-NSGA-II-GRC achieves convergence metric improvements of about 5.8% and 4.4% over the original NSGA-II, while preserving a well-distributed set of non-dominated solutions. In the portfolio application, the method produces a smooth and densely populated efficient frontier that supports the identification of the maximum Sharpe ratio portfolio (with annualized Sharpe ratio = 1.92), as well as utility-optimal portfolios for different risk-aversion levels. The main contributions of this work are three-fold: 1) we propose an RL-NSGA-II-GRC method that integrates an RL agent into the evolutionary framework to adaptively control key parameters using generational feedback; 2) we design a GRC-enhanced binary tournament selection operator that provides a comprehensive performance indicator to efficiently guide the search toward the Pareto-optimal front; 3) we demonstrate, on benchmark MOO problems and a NASDAQ portfolio case study, that the proposed method delivers improved convergence and well-populated efficient frontiers that support actionable investment insights.

## 1. Introduction

Multi-objective optimization (MOO) arises naturally in many real-world optimization and decision problems where multiple conflicting goals must be balanced simultaneously under practical constraints (Nabavi et al., 2024). In finance, the classical trade-off between risk and return in portfolio management is a canonical example, but similar structures appear in many other domains, such as engineering design, logistics, healthcare, and energy systems. Decision-makers are often interested not in a single optimal solution, but in a diverse set of Pareto-optimal solutions (i.e., non-dominated solutions or Pareto-optimal front) that reveal the trade-offs among objectives. MOO can generate this set of equally good optimal solutions from the perspective of the objectives considered (Wang et al., 2020). Designing algorithms that can efficiently approximate such Pareto-optimal fronts, especially in the presence of complex feasibility regions, remains a central challenge.

In the domain of financial application, Markowitz mean-variance framework (Zhang et al., 2018) established the foundation for quantitative portfolio selection by posing the problem as a bi-objective optimization of expected return and variance. Building on this, multi-objective evolutionary algorithms, such as the non-dominated sorting genetic algorithm II (NSGA-II) proposed by Deb et al. (2002), have become popular tools for finding the Pareto-optimal fronts (also known as efficient frontiers in finance) in portfolio optimization and other domains, due to their population-based search and ability to handle non-convex, discontinuous Pareto sets. However, when constraints are tight or the feasible region is narrow, the standard/original NSGA-II with fixed parameters and a purely feasibility or penalty-based constraint-handling strategy may suffer from slow convergence, premature loss of diversity, or excessive sampling of infeasible solutions, which reduces both the quality and interpretability of the resulting Pareto-optimal fronts.

A key factor underlying these issues is that evolutionary algorithms are typically run with static parameter settings and rigid constraint-handling rules. In practice, the balance between exploration and exploitation, or between searching in feasible versus infeasible regions, changes over the course of evolutionary generations. Early in the search, a higher degree of exploration and a more tolerant attitude toward constraint violations may be beneficial for discovering promising regions, whereas in later generations a stronger emphasis on convergence and feasibility is desirable. Fixed crossover and mutation probabilities, fixed constraint tolerance, and rigid survival selection can therefore be suboptimal, especially in constrained MOO problems where diversity and feasibility must be managed concurrently.

In parallel, recent advances in reinforcement learning (RL) and data-driven multi-criteria decision-making (MCDM) methods offer new opportunities for dynamic, feedback-driven control within evolutionary frameworks. RL provides a natural way to model the adaptive adjustment of algorithmic parameters as a sequential decision problem, where the RL agent observes indicators of convergence, feasibility, and diversity and chooses parameter values to maximize long-term performance. On the other hand, in the multi-criteria and uncertain decision-making environments, MCDM is typically an effective approach (Baydaş et al., 2025). Gray relational coefficients (GRC), a popular MCDM concept, provide a flexible mechanism for aggregating multiple indicators into a performance score that can guide selection and ranking (Wang et al., 2024). Nonetheless, the integration of RL agent with GRC-enhanced selection within NSGA-II has been relatively unexplored, particularly in the context of constrained multi-objective portfolio optimization.

To address these gaps, this work proposes a RL guided NSGA-II method enhanced with GRC (abbreviated as RL-NSGA-II-GRC) for solving MOO problems, and applies it to NASDAQ portfolio optimization. Figure 1 shows the flowchart of the RL-NSGA-II-GRC method, and the detailed description and corresponding equations are presented in Section 3 of this article.

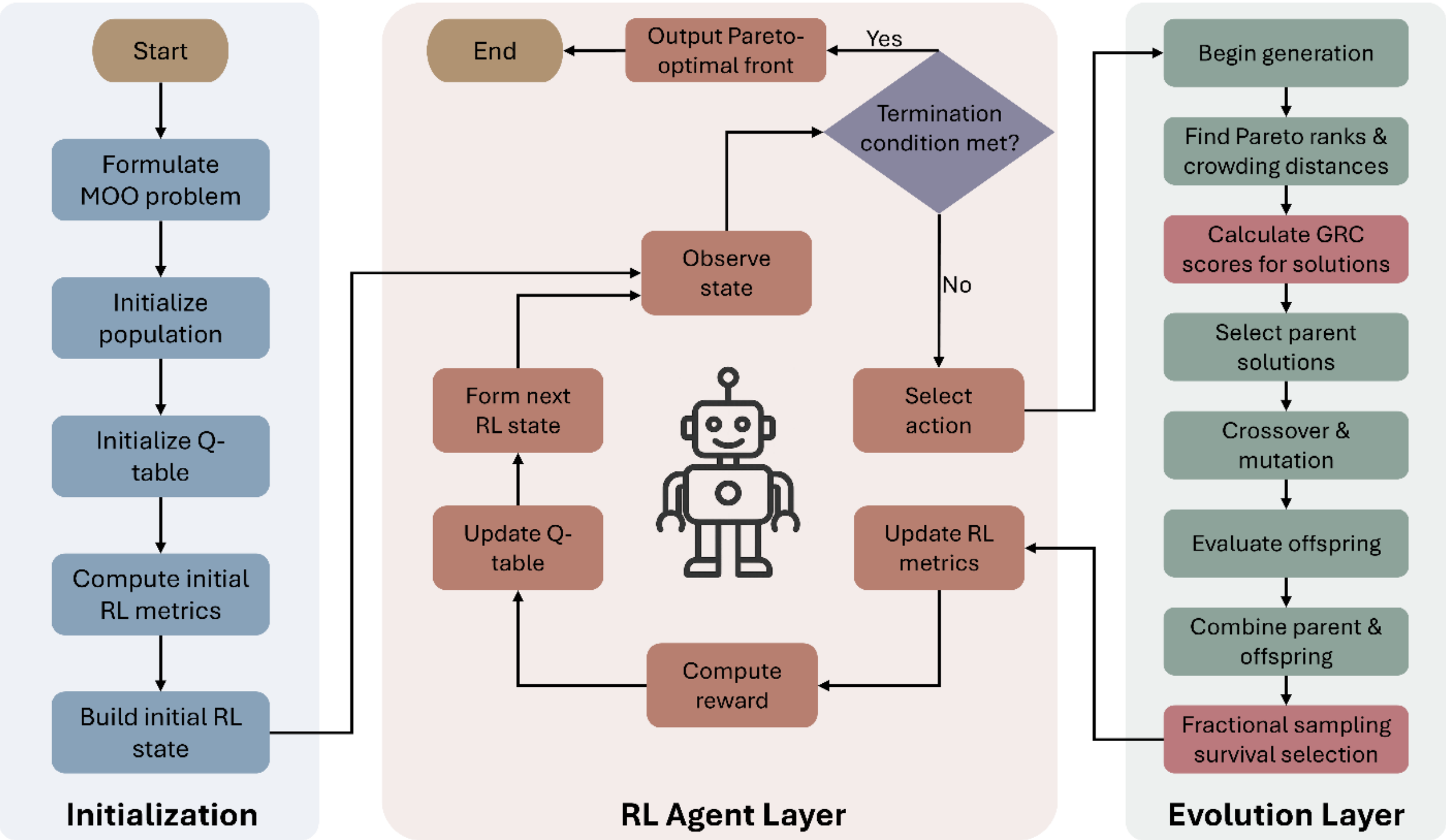

Figure 1. Flowchart of the RL-NSGA-II-GRC method.

The main contributions of this work are three-fold. First, we propose a novel RL-NSGA-II-GRC method that integrates a RL agent into the evolutionary framework to adaptively control key parameters, including crossover probability, mutation strength, constraint tolerance, and front sampling fraction; by utilizing population-level metrics (i.e., hypervolume, feasibility ratio, and diversity) as state inputs, the algorithm dynamically balances the trade-off between exploration and exploitation throughout the optimization process. Second, we introduce a GRC-enhanced binary tournament selection operator that evaluates potential parent solutions based on their geometric proximity to an ideal reference solution. This mechanism serves as a comprehensive performance indicator that simultaneously accounts for dominance rank, crowding distance, and the values of objectives, thereby improving the selection pressure and guiding the search toward the Pareto-optimal front more effectively. Third, we demonstrate the effectiveness and practicality of RL-NSGA-II-GRC on two mathematical benchmark MOO problems (i.e., Kursawe and CONSTR problems), achieving convergence improvements of about 5.8% and 4.4% over the original NSGA-II, and on a real-world NASDAQ-100 portfolio optimization case study, where the method yields a smooth and well-populated efficient frontier that supports the identification of critical investment points, such as the maximum Sharpe ratio portfolio and utility-optimal portfolios, providing actionable insights for investors with varying risk preferences.

The rest of this article is structured as follows. Section 2 reviews the relevant literature on MOO, NSGA-II, RL, GRC, and portfolio optimization. Section 3 presents the proposed RL-NSGA-II-GRC methodology in detail, including the constraint-tolerant dominance rule, GRC-enhanced binary tournament selection, RL action's performance indicators (i.e., hypervolume, feasibility ratio, and diversity), and the Q-learning control agent. Section 4 reports numerical results on the Kursawe and CONSTR benchmark MOO problems, and Section 5 presents the NASDAQ portfolio optimization case study. Section 6 discusses the findings and limitations of the proposed approach, and Section 7 concludes the paper and outlines directions for future research.

## 2. Literature Review

As aforementioned, MOO involves optimizing two or more conflicting objectives simultaneously. Instead of a single optimal solution, MOO produces a Pareto-optimal set of solutions, where no objective can be improved without worsening another (Wang et al., 2022). This is highly relevant in finance, as portfolio problems naturally have trade-offs, for instance, maximizing return and minimizing risk (Pagnoncelli et al., 2012). Early methods for MOO often included scalarization techniques (e.g., weighted sums) to combine objectives (Pereira et al., 2022), but these require subjective weight tuning and may miss certain Pareto-optimal solutions. Modern approaches also use evolutionary algorithms or other heuristics to approximate the Pareto-optimal front. Additionally, Majumder (2021) provides a comprehensive treatment of single- and multi-objective network optimization models under diverse uncertain environments, where uncertainty is represented through paradigms such as expected-value, chance-constrained, and dependent chance-constrained formulations. The study further discusses how these uncertain models can be converted into crisp equivalents and solved using both classical optimization procedures and multi-objective solution methodologies, including the global criterion method, the ε-constraint method, and evolutionary algorithms. MOO has become an essential tool across disciplines (Gunantara, 2018); beyond finance, it is applied in engineering (Sharma & Kumar, 2022), supply chain management (Jayarathna et al., 2021), healthcare (Torkayesh et al., 2021), and more. In the context of portfolio optimization, formulating it as an MOO acknowledges the inherent conflicts and seeks an efficient frontier of portfolios offering different trade-offs. For example, a bi-objective portfolio model can optimize for return and risk concurrently (Pavlou et al., 2019), yielding a spectrum of portfolios from low-risk/low-return to high-risk/high-return options. This multi-objective view, pioneered by Markowitz mean-variance framework (Zhang et al., 2018), underpins much of modern portfolio theory.

One of the most influential techniques for MOO is NSGA-II, introduced by Deb et al. (2002). NSGA-II is a widely adopted evolutionary algorithm that uses a fast non-dominated sorting approach and a crowding distance mechanism to maintain solution diversity on the Pareto front (Wang et al., 2025). First, its population is partitioned into Pareto fronts using non-dominated sorting, where solutions in the first front are non-dominated by any other solution, the second front is dominated only by members of the first front, and so on. Each solution is assigned a rank equal to its front index, where a smaller rank indicates a better Pareto status (Deb et al., 2002). Second, to maintain diversity along the Pareto front, NSGA-II computes the crowding distance within each front by sorting solutions along each objective and estimating the local density based on the distances to nearest neighbors in objective space; solutions with larger crowding distance are preferred to encourage a well-spread approximation set. Third, NSGA-II adopts an elitist replacement strategy by combining parent and offspring populations and selecting the next generation based on rank first and crowding distance second, typically implemented via a rank-and-crowding comparison in tournament selection. These mechanisms collectively yield a fast and effective baseline for MOO, with time complexity on the big O notation of $O(MN^2)$, where $M$ is the number of objectives and $N$ is the population size (Abdallah et al., 2025).

Over the last two decades, NSGA-II has attracted extensive research interest and remains one of the most commonly used methods for MOO problems (Verma et al., 2021). Its popularity stems from its efficiency and robust performance across many domains (Nabavi et al., 2025). The recent comprehensive review by Ma et al. (2023) confirms that NSGA-II and its variants are widely adopted and continue to be a cornerstone of MOO studies. In the finance literature, NSGA-II has been frequently used to solve portfolio optimization problems, generating Pareto-optimal portfolios under multiple objectives. For instance, researchers have applied NSGA-II to identify efficient frontiers when considering objectives such as return, risk, and other goals (e.g., liquidity, carbon footprint) in portfolio selection. Comparisons show that NSGA-II often performs competitively with other metaheuristics in this domain (Abdallah et al., 2025). NSGA-II elitist strategy makes it a reliable choice for tackling the multi-objective nature of portfolio optimization. Recent works continue to extend NSGA-II (e.g., with hybrid deep learning models or problem-specific improvements) to enhance its effectiveness on complex, large-scale portfolio problems (Joshi & Dhodiya, 2025). Beyond classical NSGA-II, recent studies have proposed autonomous constrained MOO frameworks that adapt search behavior based on process knowledge and population feasibility states, including the process knowledge-guided autonomous evolutionary optimization method (Zuo et al., 2023) and the population feasibility state guided autonomous constrained algorithms (Zuo & Xue, 2024), where feasibility-related state information is explicitly modeled to guide adaptive control (e.g., via learned operator decisions). These works are closely related to our study in emphasizing feasibility-aware and adaptive guidance for constrained MOO.

RL is a paradigm of machine learning where an agent learns to make sequential decisions through interactions with an environment. Formally, the problem is often modeled as a Markov decision process, and the agent's goal is to learn a policy that maximizes cumulative reward. RL differs from conventional static optimization or supervised learning in that it explicitly considers the feedback loop of decisions and outcomes over time (Rezaei & Nezamabadi-Pour, 2025). In the recent years, RL has demonstrated effectiveness across a wide range of complex decision-making problems and has attracted researchers in different research fields. From a theoretical point of view, Gu et al. (2024) surveyed the related work on the methods, theories, and applications of safe RL. Milani et al. (2024) provided a survey and comprehensive review on the explainable RL, whose objective is to explain the decision-making process of RL agents in sequential decision-making settings. Moos et al. (2022) reviewed the literature on the robust approaches to RL from four different perspectives, namely, transition robust, disturbance robust, action robust, and observation robust. Yan et al. (2024) studied a self-adapting Q-learning based inter-layer feedback mechanism in multilayer networks, where an RL agent in the management layer learns when to apply punishment to trigger game transitions and thereby promote the evolution of cooperation.

For the domain-specific RL applications, Panzer & Bender (2022) systematically reviewed the applications of deep RL in the optimization production systems, in order to overcome the challenges such as shorter product development cycles and increasing product customization. Rolf et al. (2023) reviewed the related works on RL algorithms and applications in supply chain management, including areas such as inventory management. Kayhan & Yildiz (2023) conducted a survey of the applications of RL in machine scheduling problems, such as job shop scheduling and unrelated parallel machine scheduling problems. In finance, and portfolio management particularly, RL has garnered significant attention in the past decade as a tool to assist portfolio optimization (Almahdi & Yang, 2017; Betancourt & Chen, 2021). In addition, Song et al. (2023) proposed a genetic algorithm based on RL for the electromagnetic detection satellite scheduling problem. Song et al. (2024) also conducted a comprehensive survey on RL-assisted evolutionary algorithms, proposing a taxonomy for integration schemes, analyzing RL-assisted strategies for solution generation, objective modeling, operator and parameter adaptation, and demonstrating their performance and research challenges across benchmark problems and real-world applications. Guo et al. (2025) proposed an RL assisted genetic programming algorithm for the team formation problem, in which an RL agent adaptively selects among multiple population search modes.

Introduced by Deng (1982), the gray system theory provides tools for decision-making and analysis under uncertainty. A key technique from this theory is gray relational analysis, which employs GRC to quantify the similarity or relationship between data series (Maidin et al., 2022). In essence, it measures how closely an alternative's attributes match an ideal target sequence by examining the geometric proximity of their data curves. The alternative that is closer to the reference solution receives higher ranking (Wang et al., 2023). One advantage of gray relational techniques is that they can work with small and/or incomplete data sets by extracting useful information from partial known data (Mehlawat et al., 2023). Multiple variants of gray relational analysis exist in the MCDM literature. The specific variant referenced here in this work, following Song & Jamalipour (2005) and Martinez-Morales et al. (2010), is distinguished by its independence from predefined criterion weights.

In the aspect of portfolio optimization, investors seek to maximize expected return for a given level of variance (denoting risk/uncertainty in financial management), or equivalently minimize risk for a target return, ultimately leading to an efficient frontier of optimal portfolios. The Markowitz framework formalized the risk-return trade-off and emphasized diversification benefits (i.e., considering covariances among assets) (Abdallah et al., 2025). Over time, numerous extensions and alternative formulations are proposed on top of the original framework (Markowitz, 1952). For example, Konno and Yamazaki (1991) replaced variance with mean absolute deviation (MAD) to create a linear programming model more tractable for large asset universes, while Speranza (1993) proposed mean absolute semi-deviation (MASD) focusing only on downside volatility. Rockafellar & Uryasev (2000) introduced conditional value-at-risk (CVaR) as a coherent risk measure targeting tail losses, which has since been widely adopted in portfolio optimization. These developments reflect a broadening of risk measures. aligning the optimization with investors' true risk concerns, that is, avoiding large losses. Traditional quadratic programming or other exact solvers struggle with solving such complex real-world optimization problems, especially when multiple objectives are present (Rezaei & Nezamabadi-Pour, 2025). To tackle this, researchers turned to metaheuristics and artificial intelligence (AI)-based methods. Global search algorithms like NSGA-II and other evolutionary algorithms have been extensively applied to portfolio problems with promising results (Abdallah et al., 2025). These methods can accommodate non-linear, discontinuous objective landscapes and find near-optimal solutions under complex constraints (Zhan et al., 2022). For instance, there is evidence that evolutionary algorithms and swarm intelligence methods can efficiently construct diversified portfolios that classical solvers might miss (Zhu et al., 2011). Anagnostopoulos & Mamanis (2011) found that an evolutionary approach outperformed exact solvers on a constrained portfolio selection benchmark, and many subsequent studies have confirmed the utility of such heuristics for multi-objective portfolio optimization (Erwin & Engelbrecht, 2023). Recently, the learnheuristic (Bullah & van Zyl, 2023) or hybrid methods have emerged, which combine machine learning models with metaheuristics optimization algorithms. For example, Joshi & Dhodiya (2025) recently proposed a hybrid deep learning and evolutionary algorithm framework for many-objective portfolio decisions, illustrating the trend of blending AI techniques with metaheuristics algorithms.

Based on our comprehensive literature review, although RL-assisted multi-objective evolutionary algorithms (MOEAs) have been increasingly studied, a substantial portion of existing efforts mainly focuses on operator/parameter control (e.g., adapting crossover/mutation rates) or selecting among predefined search modes, while keeping the feasibility logic and survival selection mechanism largely unchanged. In contrast, the proposed RL-NSGA-II-GRC framework differs fundamentally in that the RL agent intervenes not only in variation parameters but also in constraint handling and survival selection pressure within the NSGA-II backbone. Specifically, the RL agent jointly controls (i) the variation parameters, (ii) a generation-wise constraint tolerance that defines a constraint-tolerant dominance relation for constrained MOO, and (iii) a fractional front-sampling survival mechanism that regulates selection pressure across Pareto fronts. Moreover, rather than replacing Pareto dominance with a single scalar fitness, the proposed method preserves a Pareto-first ranking logic and introduces the GRC as a discriminator when dominance information is insufficient, thereby maintaining the key semantics of non-dominated sorting while improving decision consistency during selection.

**3. Methodology**

This section presents the proposed RL-NSGA-II-GRC method in a general MOO setting, as illustrated in the flowchart in Figure 1. For clarity, we first formulate the generic MOO problem, followed by a comprehensive description of each methodological component. Specifically, we detail: 1) the fundamental operations of NSGA-II, such as the mathematical definitions of simulated binary crossover, polynomial mutation, non-dominated sorting, and crowding distance calculation; 2) the integration of GRC into the tournament selection operator, enabling selection pressure that always simultaneously accounts for dominance rank, crowding distance, and geometric proximity to ideal objective values; 3) the computation of the population-level performance indicators (including hypervolume, feasibility ratio, and diversity) that serve as quantitative feedback on optimization progress (via calculating the reward of the RL agent's action) and are used as state inputs for the RL agent; 4) the design of the RL control layer which adaptively takes action and adjusts evolutionary parameters (e.g., crossover probability, mutation strength, constraint tolerance, and front sampling ratio) based on learned value estimates from the environment, such that it gradually improves the algorithm's exploration and exploitation balance throughout the evolutionary process for finding the Pareto-optimal solutions (i.e., Pareto-optimal front). The complete RL-NSGA-II-GRC method is implemented in Python. The code is available at no cost to interested readers by contacting the corresponding authors of this paper.

We consider a generic MOO problem involving two or more objective functions to be minimized (i.e., $f_1, f_2, \dots, f_M$). Note that any maximization objective can be readily converted to a minimization form by multiplying its objective function by $-1$ (negative one). The problem is defined over a decision variables vector ($\mathbf{x}$), subject to lower and upper bounds on each decision variable, as well as a set of inequality and/or equality constraints. The mathematical formulation of MOO is as follows:

$$\min_{\mathbf{x} \in \Omega} \mathbf{F}(\mathbf{x}) = \big(f_1(\mathbf{x}), f_2(\mathbf{x}), \dots, f_M(\mathbf{x})\big), \tag{1}$$

subject to inequality constraints

$$g_j(\mathbf{x}) \geq 0, \quad j = 1,2, \dots, J_g, \tag{2}$$

and/or equality constraints

$$h_\ell(\mathbf{x}) = 0, \quad \ell = 1,2, \dots, J_h, \tag{3}$$

and bound constraints

$$x_k^{(L)} \le x_k \le x_k^{(U)}, \quad k = 1,2,\dots,D, \tag{4}$$

where $\mathbf{x} = (x_1, x_2, \dots, x_k \dots, x_D)^\top$ is a decision vector in the search space $\Omega \subseteq \mathbb{R}^D$, and $\mathbf{F}(\mathbf{x}) \in \mathbb{R}^M$ collects the total $M$ objective functions to be minimized.

To quantify constraint violation for a solution $\mathbf{x}$, we use the following scalar constraint violation

$$\mathrm{CV}(\mathbf{x}) = \sum_{j=1}^{J_g} \max\left(0, -g_j(\mathbf{x})\right) + \sum_{\ell=1}^{J_h} |h_\ell(\mathbf{x})|\,. \tag{5}$$

Thus, $\mathrm{CV}(\mathbf{x}) = 0$ if and only if all constraints are satisfied, with larger values indicating greater constraint violation.

For dominance, a solution $\mathbf{x}$ is said to Pareto-dominate another solution $\mathbf{y}$ (denoted $\mathbf{x} \prec \mathbf{y}$) if and only if

$$\begin{aligned} &f_m(\mathbf{x}) \le f_m(\mathbf{y}) \quad \forall m = 1,2,\dots,M, \\ &\exists\, m: f_m(\mathbf{x}) < f_m(\mathbf{y}). \end{aligned} \tag{6}$$

In other words, $\mathbf{x}$ Pareto-dominates $\mathbf{y}$ if $\mathbf{x}$ is no worse than $\mathbf{y}$ in all $M$ objectives, and at the same $\mathbf{x}$ is strictly better than $\mathbf{y}$ in at least one of the objectives.

### 3.1. NSGA-II backbone with dynamic constraint-tolerant dominance and fractional sampling survival selection

During the evolutionary process of NSGA-II, each individual solution in the population corresponds to a decision vector $\mathbf{x}$, augmented with objective function vector $\mathbf{F}(\mathbf{x})$, constraint violation $\mathrm{CV}(\mathbf{x})$, Pareto rank (e.g., rank-0, rank-1), crowding distance that measures sparsity, and additional bookkeeping for dominance (i.e., domination count and dominated solutions set).

#### 3.1.1. Constraint-tolerant dominance

In classical original NSGA-II, the rules of Deb et al. (2002) enforce strict feasibility: any feasible solution is preferred over any infeasible one; if both are infeasible, the smaller violation is preferred. In our proposed RL-NSGA-II-GRC method, this is generalized by introducing a dynamic generation-dependent constraint tolerance $\tau_t \geq 0$, chosen by the RL agent at generation $t$.

For an individual solution $p$ with violation $\mathrm{CV}(p)$, we define the effective violation:

$$\widetilde{\mathrm{CV}}_t(p) = \begin{cases} 0, & \text{if } \mathrm{CV}(p) \leq \tau_t, \\ \mathrm{CV}(p), & \text{otherwise.} \end{cases} \tag{7}$$

Any solution whose violation is less than or equal to $\tau_t$ is treated as effectively feasible in generation $t$. This allows certain degree of controlled relaxation of constraints to help explore across the infeasible regions. Theoretically, $\tau_t$ is defined as a non-negative relaxation threshold, i.e., $\tau_t \geq 0$. In practice, $\tau_t$ is applied as a small tolerance level to determine whether a solution is treated as effectively feasible when computing $\widetilde{\mathrm{CV}}_t(p)$. Accordingly, $\tau_t = 0$ corresponds to the standard strict feasibility case (i.e., no relaxation), while larger $\tau_t$ values allow controlled relaxation by treating violations up to $\tau_t$ as acceptable during search. In our Python code implementation, $\tau_t$ is decided adaptively by the RL agent controller with value up to 0.5 (this is also tunable), so that the relaxation level remains bounded.

Given two individual solutions $p$ and $q$, the constraint-tolerant dominance is therefore defined in the following three scenarios. This relation $\prec_t$ is used in all dominance-based operations at generation $t$.

First, if one solution is effectively feasible and the other is not, then the effectively feasible solution dominates:

$$\text{If } \widetilde{\mathrm{CV}}_t(p) = 0 \text{ and } \widetilde{\mathrm{CV}}_t(q) > 0, \qquad \text{then } p \prec_t q. \tag{8}$$

Second, if both solutions are effectively infeasible, then the solution with the smaller effective violation is preferred:

$$\text{If } \widetilde{\mathrm{CV}}_t(p) > 0,\ \widetilde{\mathrm{CV}}_t(q) > 0, \text{ and } \widetilde{\mathrm{CV}}_t(p) < \widetilde{\mathrm{CV}}_t(q), \qquad \text{then } p \prec_t q \tag{9}$$

Third, if both solutions satisfy the tolerance, that is, $\widetilde{\mathrm{CV}}_t(p) = \widetilde{\mathrm{CV}}_t(q) = 0$, then the dominance determination reduces to the standard Pareto dominance rule as in Eq (6).

#### 3.1.2. Non-dominated sorting

Given a population $\mathcal{P}_t$ of size $N$, non-dominated sorting partitions it into fronts $\mathcal{F}_0, \mathcal{F}_1, \ldots$ with increasing rank. For each individual $p \in \mathcal{P}_t$, we define:

the set of solutions it dominates

$$S_p = \{q \in \mathcal{P}_t \mid p \prec_t q\}, \tag{10}$$

and the number of solutions that dominate it

$$n_p = |\{ q \in \mathcal{P}_t \mid q \prec_t p \}|. \tag{11}$$

Here, the first front (rank-0) is:

$$\mathcal{F}_0 = \left\{p \in \mathcal{P}_t \mid n_p = 0\right\}, \quad \mathrm{rank}(p) = 0. \tag{12}$$

Subsequent fronts are built iteratively:

$$\mathcal{F}_{\ell+1} = \left\{q \;\middle|\; \exists p \in \mathcal{F}_\ell\colon\ q \in S_p,\ n_q \text{ decremented to } 0\right\}, \quad \ell = 0,1, \ldots \tag{13}$$

and individuals in $\mathcal{F}_{\ell+1}$ are assigned $\mathrm{rank}(q) = \ell + 1$. The process continues until all individuals are assigned to a front.

#### 3.1.3. Crowding distance calculation

Within each front $\mathcal{F}_\ell$, a crowding distance is computed to measure the sparsity of solutions in objective space and to encourage a well-spread Pareto front. Let $|\mathcal{F}_\ell| = K$. Initialize all distances:

$$d(p) = 0, \quad \forall p \in \mathcal{F}_\ell. \tag{14}$$

Then, for each objective $m = 1, \ldots, M$, sort the front with respect to objective $m$:

$$\mathcal{F}_\ell^{(m)} = \mathrm{sort}(\mathcal{F}_\ell, \text{ by } f_m). \tag{15}$$

The boundary points are assigned infinite distance:

$$d\left(\mathcal{F}_\ell^{(m)}[1]\right) = d\left(\mathcal{F}_\ell^{(m)}[K]\right) = +\infty. \tag{16}$$

For internal points $i = 2, \dots, K-1$, the contribution of objective $m$ to the crowding distance is:

$$d^{(m)}\left(\mathcal{F}_\ell^{(m)}[i]\right) = \frac{f_m\left(\mathcal{F}_\ell^{(m)}[i+1]\right) - f_m\left(\mathcal{F}_\ell^{(m)}[i-1]\right)}{f_m\left(\mathcal{F}_\ell^{(m)}[K]\right) - f_m\left(\mathcal{F}_\ell^{(m)}[1]\right)}. \tag{17}$$

The total crowding distance for individual $p$ is the sum across $M$ objectives:

$$d(p) = \sum_{m=1}^{M} d^{(m)}\,(p). \tag{18}$$

A larger $d(p)$ indicates that the solution $p$ is located in a sparsely populated region of the front and thus should be preferred for diversity.

#### 3.1.4. Simulated binary crossover (SBX)

The NSGA-II uses crossover and mutation to generate offspring from selected parents. The key steps of the simulated binary crossover (SBX) are briefly presented as follows. Given two parents $\mathbf{x}^{(1)}$ and $\mathbf{x}^{(2)}$, SBX simulates a recombination similar to single-point crossover in binary representation, but in real variables. Let the crossover probability be $p_c$ and the distribution index be $\eta_c > 0$. Both the $p_c$ and $\eta_c$ are to be dynamically controlled by the RL agent in our RL-NSGA-II-GRC method.

Here, for each decision variable $j = 1, \dots, D$, with probability $1 - p_c$, the parents' genes (i.e., parents' decision variable values) are directly copied to offspring:

$$c_j^{(1)} = x_j^{(1)}, \quad c_j^{(2)} = x_j^{(2)}. \tag{19}$$

Otherwise, SBX is performed,

$$y_1 = \min\left(x_j^{(1)}, x_j^{(2)}\right), \quad y_2 = \max\left(x_j^{(1)}, x_j^{(2)}\right), \tag{20}$$

and the lower and upper bounds of the decision variable $j$ still apply as in Eq. (4), that is, $[L_j, U_j]$ bounds. Then, uniformly draw a random number $u \sim \mathcal{U}(0,1)$, and compute:

$$\beta_1 = 1 + \frac{2(y_1 - L_j)}{y_2 - y_1}, \quad \alpha_1 = 2 - \beta_1^{-(\eta_c+1)}, \tag{21}$$

$$\beta_q = \begin{cases} (u\alpha_1)^{\frac{1}{(\eta_c+1)}}, & u \le \frac{1}{\alpha_1}, \\ \left(\frac{1}{2 - u\alpha_1}\right)^{\frac{1}{(\eta_c+1)}}, & u > \frac{1}{\alpha_1}. \end{cases} \quad (22)$$

The first child gene for the $j^{th}$ variable is obtained as:

$$c_j^{(1)} = \frac{1}{2}\left[(y_1 + y_2) - \beta_q(y_2 - y_1)\right]. \quad (23)$$

On the other side of the interval,

$$\beta_2 = 1 + \frac{2(U_j - y_2)}{y_2 - y_1}, \quad \alpha_2 = 2 - \beta_2^{-(\eta_c+1)}, \quad (24)$$

and with the same $u$,

$$\beta_q = \begin{cases} (u\alpha_2)^{\frac{1}{(\eta_c+1)}}, & u \le \frac{1}{\alpha_2}, \\ \left(\frac{1}{2 - u\alpha_2}\right)^{\frac{1}{(\eta_c+1)}}, & u > \frac{1}{\alpha_2}. \end{cases} \quad (25)$$

The second child gene for the $j^{th}$ variable is obtained as:

$$c_j^{(2)} = \frac{1}{2}\left[(y_1 + y_2) + \beta_q(y_2 - y_1)\right]. \quad (26)$$

Both $c_j^{(1)}$ and $c_j^{(2)}$ are finally clipped to their bounds:

$$c_j^{(k)} \leftarrow \min\left(\max\left(c_j^{(k)}, L_j\right), U_j\right), \quad k = 1,2. \quad (27)$$

### 3.1.5. Polynomial mutation

Given a mutation probability $p_m$ per variable and distribution index $\eta_m$, polynomial mutation perturbs each variable $x_j$ with probability $p_m$. Likewise, both $p_m$ and $\eta_m$ are dynamically controlled by the RL agent in our proposed RL-NSGA-II-GRC method. For a selected variable $x_j$, define:

$$\delta_1 = \frac{x_j - L_j}{U_j - L_j}, \quad \delta_2 = \frac{U_j - x_j}{U_j - L_j}, \quad (28)$$

and uniformly draw a random number $u \sim \mathcal{U}(0,1)$. Let $m = \frac{1}{\eta_m+1}$. The mutation step $\delta_q$ is given by:

$$\delta_q = \begin{cases} (2u + (1-2u)(1-\delta_1)^{\eta_m+1})^{\frac{1}{\eta_m+1}} - 1, & u \le 0.5, \\ 1 - (2(1-u) + 2(u-0.5)(1-\delta_2)^{\eta_m+1})^{\frac{1}{\eta_m+1}}, & u > 0.5 \end{cases} \tag{29}$$

The mutated value at the $j^{th}$ variable is:

$$x'_j = x_j + \delta_q(U_j - L_j), \tag{30}$$

followed by bound clipping:

$$x'_j \leftarrow \min(\max(x'_j, L_j), U_j). \tag{31}$$

**3.2. Gray relational coefficient (GRC) enhanced binary tournament selection**

To enhance parent selection beyond Pareto rank and crowding distance alone, RL-NSGA-II-GRC employes a MCDM approach to derive the gray relational coefficient (GRC) score for each potential parent solution, which combines the performances of objectives and crowding distance information in a normalized manner. GRC is founded on the gray system theory. This theory serves as a powerful analytical tool, particularly in situations where information is incomplete or uncertain. The core principle of GRC is to evaluate solutions by measuring their proximity to an ideal reference solution (Wang & Rangaiah, 2025). The solution that is closer to the ideal solution receives higher GRC score.

At generation $t$, consider the current population $\mathcal{P}_t$. For each objective $m = 1, \dots, M$, compute the population-level extremes:

$$f_m^{\min} = \min_{p \in \mathcal{P}_t} f_m(p), \quad f_m^{\max} = \max_{p \in \mathcal{P}_t} f_m(p). \tag{32}$$

Because all objectives are minimized by default, we normalize each objective so that the value 1 represents the best performance, and the value 0 denotes the worst performance consistently.

$$\hat{f}_m(p) = \begin{cases} \dfrac{f_m^{\max} - f_m(p)}{f_m^{\max} - f_m^{\min}}, & f_m^{\max} \neq f_m^{\min}, \\ 1, & \text{otherwise}, \end{cases} \quad \hat{f}_m(p) \in [0,1]. \tag{33}$$

For crowding distance, recall that we have already calculated the crowding distance of solution $p$ (i.e., $d(p)$) in the previous section using Eq. (18), and the boundary solutions have $d(p) = +\infty$. Here, we further compute

$$d^{\min} = \min_{p \in \mathcal{P}_t,\, d(p) < +\infty} d(p), \quad d^{\max} = \max_{p \in \mathcal{P}_t,\, d(p) < +\infty} d(p). \tag{34}$$

Infinite distances (for boundary solutions) are replaced by a finite proxy to facilitate the subsequent calculations.

$$D^{\max} = \begin{cases} 1.2\, d^{\max}, & d^{\max} > 0, \\ 1.0, & d^{\max} = 0. \end{cases} \tag{35}$$

Then the normalized crowding distance is:

$$\hat{d}(p) = \begin{cases} \dfrac{d(p) - d^{\min}}{D^{\max} - d^{\min}}, & D^{\max} \neq d^{\min}, \\ 1, & \text{otherwise}. \end{cases} \quad \hat{d}(p) \in [0,1]. \tag{36}$$

Next, each normalized criterion $\hat{y} \in [0,1]$ (either an objective or crowding distance) is compared to their ideal reference value $\hat{y}_{\text{ref}} = 1$. The GRC for that criterion is:

$$\xi(\hat{y}) = \frac{1}{|\hat{y}_{\text{ref}} - \hat{y}| + 1} = \frac{1}{|1 - \hat{y}| + 1}, \tag{37}$$

where $\xi(\hat{y})$ indicates that $\hat{y}$ is closer to the ideal. For each individual $p$, we compute one coefficient for the crowding distance:

$$\xi_d(p) = \frac{1}{|1 - \hat{d}(p)| + 1}, \tag{38}$$

and one coefficient for each objective $m$:

$$\xi_m(p) = \frac{1}{|1 - \hat{f}_m(p)| + 1}. \tag{39}$$

The overall GRC score is defined as the average of these $M + 1$ coefficients:

$$\text{GRC}(p) = \frac{1}{M+1}\left(\xi_d(p) + \sum_{m=1}^{M} \xi_m(p)\right), \tag{40}$$

where solution $p$ with larger $\text{GRC}(p)$ score is preferred.

After finding the GRC scores for solutions, the binary tournament selection proceeds straightforwardly. Given two randomly sampled individuals $p$ and $q$, if $\text{rank}(p) < \text{rank}(q)$, we select $p$; else if $\text{rank}(q) < \text{rank}(p)$, we select $q$; otherwise (means same Pareto rank, which common occurs especially toward the later stage of evolution), we compare their GRC scores.

$$\text{select} \begin{cases} p, & \text{if } \text{GRC}(p) > \text{GRC}(q), \\ q, & \text{otherwise.} \end{cases} \tag{41}$$

In this way, the GRC score serves as a scalar indicator balancing objective quality and spread of solutions.

#### 3.2.1. Survival selection with fractional sampling mechanism

After parent selection, crossover, and mutation steps, we are able to obtain the set of new offspring solutions $\mathcal{Q}_t$, the parent and offspring solutions are merged into a combined pool $\mathcal{R}_t = \mathcal{P}_t \cup \mathcal{Q}_t$ of size $2N$ (where $N$ is the predefined number of solutions desired in the population). This pool is first partitioned into non-dominated fronts $\{\mathcal{F}_0(t), \mathcal{F}_1(t), \ldots\}$ using the constraint-tolerant dominance relation $\prec_t$ (as discussed in Sections 3.11 and 3.12). Within each front $\mathcal{F}_\ell(t)$, crowding distances are computed to promote a good spread in objective space. Unlike classical NSGA-II, which always admits the entire $\mathcal{F}_0$ and subsequent fronts until the population budget is nearly exhausted, our proposed RL-NSGA-II-GRC method adopts a fractional front sampling mechanism dynamically controlled by the RL agent. Specifically, for each front $\mathcal{F}_\ell(t)$, only

$$n_\ell^*(t) = \min\left(N - |\mathcal{P}_{t+1}|, \max(1, \lfloor \varphi_t \rfloor \, |\mathcal{F}_\ell(t)|)\right) \tag{42}$$

solutions are selected, where $\varphi_t \in (0,1]$ is the front sampling fraction managed by the RL agent at generation $t$ based on the environment. The selected individuals are the $n_\ell^*(t)$ solutions of $\mathcal{F}_\ell(t)$ with the largest crowding distance, ensuring that the most diverse solutions from each front are retained. If, after scanning all fronts, the size of the new population $\mathcal{P}_{t+1}$ is still smaller than the desired population $N$, the remaining slots are filled from the leftover individuals in $(\mathcal{R}_t \setminus \mathcal{P}_{t+1})$, sorted lexicographically by (rank, $-$crowding distance).

This two-stage survival selection scheme departs from the behavior of standard NSGA-II: by allowing $\varphi_t < 1$ (e.g., $\varphi_t = 0.9$), the algorithm may deliberately not take all members of $\mathcal{F}_0(t)$, thereby leaving room for selected individuals from some inferior fronts to survive in order to boost diversity (and also enhance the different possibilities of offspring solutions in next generation). The degree of this relaxation is adaptively controlled by the RL agent based on the observed hypervolume, feasibility ratio, and diversity metrics (discussed in Section 3.3) in the environment. As a result, the survival selection becomes a dynamic balance between exploitation of the current best front and exploration via lower-rank but diverse or promising solutions, which can be particularly beneficial on constrained MOO problems.

### 3.3. Population-level performance indicators for reinforcement learning (RL)

Three population-level indicators, namely, hypervolume, feasibility ratio, and diversity are utilized to evaluate the impact of RL agent's actions. These are computed at every evolutionary generation $t$.

#### 3.3.1. Hypervolume

Let $\mathcal{F}_0^{\text{feas}}(t)$ denote the feasible rank-0 front (i.e., a set of the best solutions) at generation $t$:

$$\mathcal{F}_0^{\text{feas}}(t) = \{p \in \mathcal{F}_0(t) \mid \text{CV}(p) = 0\}. \quad (43)$$

Let $\mathbf{r} \in \mathbb{R}^M$ be a fixed reference point chosen to be dominated by all relevant objectives (e.g., constructed from early population extremes with a safety margin). The hypervolume (HV) at generation $t$ is the Lebesgue measure of the union of objective-space hyperrectangles dominated by $\mathcal{F}_0^{\text{feas}}(t)$ and bounded by $\mathbf{r}$:

$$\text{HV}_t = V_L\left(\bigcup_{p \in \mathcal{F}_0^{\text{feas}}(t)} \{\mathbf{z} \in \mathbb{R}^M | f_m(p) \leq z_m \leq r_m,\ m = 1,2,\dots,M\}\right), \tag{44}$$

where $V_L(\cdot)$ denotes the $M$-dimensional Lebesgue measure of volume. For $M = 2$, this basically reduces to the area dominated by the rank-0 front. A larger hypervolume value is preferred as it indicates a Pareto front that is closer to the true optimum.

### 3.3.2. Feasibility ratio

The feasibility ratio at generation $t$ is the fraction of the current population ($\mathcal{P}_t$) with zero constraint violation:

$$\text{FR}_t = \frac{1}{N} \sum_{p \in \mathcal{P}_t} \mathbb{I}\,(\text{CV}(p) = 0), \tag{45}$$

where $\mathbb{I}(\cdot)$ is the indicator function, which returns $\mathbb{I}(\text{True}) = 1$ and $\mathbb{I}(\text{False}) = 0$. As such, each feasible solution contributes 1 to the sum, while each infeasible solution contributes 0. Consequently, $\text{FR}_t \in [0,1]$ quantifies the feasibility level of the current population, with values closer to 1 being preferred, indicating that the evolutionary search successfully maintains feasibility across generations.

### 3.3.3. Diversity

The diversity metric measures how well the population is spread across the objective space, reflecting the algorithm's ability to explore multiple trade-offs rather than just clustering around a narrow region. A larger diversity value indicates a more widely dispersed set of solutions, which is desirable because it supports the discovery of a well-distributed Pareto front.

To obtain the diversity value for rank-0 front ($p \in \mathcal{F}_0(t)$), we normalize all objectives to $[0,1]$ using population-level minimum and maximum values:

$$\tilde{f}_m(p) = \begin{cases} \dfrac{f_m(p) - f_m^{\min}}{f_m^{\max} - f_m^{\min}}, & f_m^{\max} \neq f_m^{\min}, \\ 0.5, & \text{otherwise}, \end{cases} \quad m = 1,2,\dots,M. \tag{46}$$

Define the normalized objective vector:

$$\tilde{\mathbf{f}}(p) = (\tilde{f}_1(p), \dots, \tilde{f}_M(p))^\top, \tag{47}$$

Let $K = |\mathcal{F}_0(t)|$ denotes the number of solutions within the rank-0 front at generation $t$. The diversity is then taken as the average pairwise Euclidean distance:

$$\mathrm{DIV}_t = \begin{cases} 0, & K < 2, \\ \dfrac{2}{K(K-1)} \displaystyle\sum_{i=1}^{K-1} \sum_{j=i+1}^{K} \left\| \tilde{\mathbf{f}}(p_i) - \tilde{\mathbf{f}}(p_j) \right\|_2, & \text{otherwise.} \end{cases} \tag{48}$$

**3.4. Reinforcement learning (RL) layer**

The RL layer views each generation as a time step in a Markov decision process. At each generation $t$, the agent observes a state summarizing the population, selects an action that sets the evolutionary parameters, and receives a scalar reward based on improvements in hypervolume, feasibility, and diversity. The state at generation $t$ is defined as:

$$s_t = \left(\sigma_t^{\mathrm{HV}}, b_t^{\mathrm{FR}}, b_t^{\mathrm{DIV}}, \phi_t\right), \tag{49}$$

which comprising the hypervolume trend sign:

$$\sigma_t^{\mathrm{HV}} = \mathrm{sign}_\epsilon(\mathrm{HV}_t - \mathrm{HV}_{t-1}), \tag{50}$$

where (1 improvement, $-1$ deterioration, 0 no change)

$$\mathrm{sign}_\epsilon(x) = \begin{cases} 1, & x > \epsilon, \\ -1, & x < -\epsilon, \\ 0, & |x| \leq \epsilon, \end{cases} \tag{51}$$

with $\epsilon > 0$, a small tolerance (e.g., $1 \times 10^{-8}$); then, feasibility level via discretized feasibility ratio (0 low, 1 medium, 2 high):

$$b_t^{\mathrm{FR}} = \begin{cases} 0, & \mathrm{FR}_t < 1/3\,, \\ 1, & 1/3 \le \mathrm{FR}_t < 2/3\,, \\ 2, & \mathrm{FR}_t \ge 2/3\,, \end{cases} \tag{52}$$

followed by the diversity level (0 low, 1 medium, 2 high):

$$b_t^{\mathrm{DIV}} = \begin{cases} 0, & \mathrm{DIV}_t < 0.2, \\ 1, & 0.2 \le \mathrm{DIV}_t < 0.4, \\ 2, & \mathrm{DIV}_t \ge 0.4, \end{cases} \tag{53}$$

and the search stage $\phi_t$, encoding whether we are in the early (0), middle (1), or late (2) stage of the run, given total generation budget $T$ (e.g., 300 generations):

$$\phi_t = \begin{cases} 0, & t < T/3, \\ 1, & T/3 \le t < 2T/3, \\ 2, & t \ge 2T/3. \end{cases} \tag{54}$$

This yields a compact, fully discrete state representation for the RL agent. For instance, a state $s_t = (1,\ 2,\ 1,\ 0)$ means an improved hypervolume (relative to the previous generation), a high feasibility level for solutions in the current population, a medium diversity level in the current rank-0 front, and an early stage of the evolutionary search.

Next, the action space $\mathcal{A} = \{a^0, \dots, a^{K-1}\}$ is a finite set of parameter modes. Each action $a^k$ encodes the following key parameters, namely, crossover probability $p_c^{(k)}$, mutation probability $p_m^{(k)}$, SBX distribution index $\eta_c^{(k)}$, mutation distribution index $\eta_m^{(k)}$, constraint tolerance $\tau^{(k)}$, and front sampling fraction $\varphi^{(k)}$ in survival selection. Formally,

$$a^k = \left(p_c^{(k)},\ p_m^{(k)},\ \eta_c^{(k)},\ \eta_m^{(k)},\ \tau^{(k)},\ \varphi^{(k)}\right). \tag{55}$$

When action $a_t$ is selected at generation $t$, the NSGA-II parameters are set as: $p_c = p_c^{(a_t)}$, $p_m = p_m^{(a_t)}, \eta_c = \eta_c^{(a_t)}, \eta_m = \eta_m^{(a_t)}, \tau_t = \tau^{(a_t)}$, and $\varphi_t = \varphi^{(a_t)}$. Here, $\tau_t$ enters the constraint-tolerant dominance, and $\varphi_t \in (0,1]$ controls how many individuals are accepted from each front in survival selection, adjusting selection pressure and diversity.

A tabular action-value function $Q(s, a)$ is maintained as a dictionary (i.e., a Q-learning table) mapping each visited state $s$ to a vector of Q-values $\{Q(s, a^k)\}_{k=0}^{K-1}$. Therefore, at the generation $t$, given state $s_t$, an ε-greedy policy selects:

$$a_t = \begin{cases} \text{rand}(\mathcal{A}), & \text{with probability } \varepsilon_t, \\ \underset{a \in \mathcal{A}}{\text{argmax}}\, Q(s_t, a), & \text{with probability } 1 - \varepsilon_t, \end{cases} \tag{56}$$

where ties in the argmax are broken uniformly at random. The exploration parameter $\varepsilon_t$ (e.g., a value 0.3) is gradually decayed:

$$\varepsilon_{t+1} = \max(\varepsilon_{\min}, \lambda \varepsilon_t), \quad 0 < \lambda < 1, \tag{57}$$

so that the policy moves from exploration to exploitation over time; for example, $\varepsilon_{\min}$ can be small value (e.g., 0.05) and $\lambda$ takes 0.995.

As seen, the discretization of the RL state space is motivated by the use of tabular Q-learning, where the number of state-action pairs must remain sufficiently small to ensure stable learning under a limited generational budget. In this study, the state is therefore designed as a compact summary of the population dynamics using (i) the sign of hypervolume change (improved/unchanged/deteriorated), (ii) discretized feasibility ratio (low/medium/high), (iii) discretized diversity level (low/medium/high), and (iv) the search stage (early/middle/late). This design is not intended to represent every population statistic; rather, it provides a parsimonious, decision-relevant representation that captures the three core evolutionary goals, namely convergence, feasibility, and spread, while also accounting for early exploration or late exploitation. Coarser discretization also reduces noise sensitivity and avoids overfitting the control policy to small metric fluctuations, which is particularly important when the same controller is expected to generalize across different constrained MOO instances.

After applying action $a_t$ and completing one evolutionary generation to get new set of population solutions, the RL agent observes newly updated metrics $\text{HV}_{t+1}$, $\text{FR}_{t+1}$, and $\mathrm{DIV}_{t+1}$. To make the reward scale-invariant, we compute the normalized gains. For hypervolume gain:

$$\Delta\mathrm{HV}_{t+1}^{\mathrm{norm}} = \frac{\mathrm{HV}_{t+1} - \mathrm{HV}_t}{|\mathrm{HV}_t|}, \tag{58}$$

and for feasibility gain:

$$\Delta\mathrm{FR}_{t+1} = \mathrm{FR}_{t+1} - \mathrm{FR}_t, \tag{59}$$

then diversity gain:

$$\Delta\mathrm{DIV}_{t+1}^{\mathrm{norm}} = \frac{\mathrm{DIV}_{t+1} - \mathrm{DIV}_t}{|\mathrm{DIV}_t|}. \tag{60}$$

The normalized quantities are typically clamped to $[-1,1]$ to stabilize learning. The immediate reward ($r_t$) is then defined as a weighted sum of these three gains:

$$r_t = w_{\mathrm{HV}}\Delta\mathrm{HV}_{t+1}^{\mathrm{norm}} + w_{\mathrm{FR}}\Delta\mathrm{FR}_{t+1} + w_{\mathrm{DIV}}\Delta\mathrm{DIV}_{t+1}^{\mathrm{norm}}, \tag{61}$$

with $w_{\mathrm{HV}} + w_{\mathrm{FR}} + w_{\mathrm{DIV}} = 1$, e.g., $w_{\mathrm{HV}} = 0.6$, $w_{\mathrm{FR}} = 0.2$, and $w_{\mathrm{DIV}} = 0.2$. This emphasizes hypervolume improvement while still rewarding better feasibility and diversity.

The reward is defined as a weighted sum of normalized gains in hypervolume, feasibility, and diversity to align the RL objective with standard performance criteria in constrained MOO. The higher weight assigned to hypervolume reflects its role as a unified indicator that simultaneously captures convergence and coverage of the Pareto front, whereas feasibility ratio and diversity are included to prevent degenerate behaviors (e.g., rapidly improving hypervolume by concentrating on infeasible regions or by collapsing diversity). The chosen weights therefore express the intended priority, that is, driving Pareto-front quality while maintaining feasibility and solution spread as necessary constraints on the search behavior. Importantly, feasibility and diversity retain nonzero weights so that the RL controller is explicitly discouraged from sacrificing feasibility and spread to obtain short-term hypervolume gains.

Finally, let $\alpha \in (0,1]$ denote the learning rate, and $\gamma \in [0,1)$ the discount factor. After observing next state $s_{t+1}$ and reward $r_t$, the temporal difference (TD) target is:

$$\mathrm{TD}_{\mathrm{target}_t} = r_t + \gamma \max_{a' \in \mathcal{A}} Q(s_{t+1}, a'), \tag{62}$$

where $r_t$ is the immediate reward, and the second term, $\gamma \max_{a' \in \mathcal{A}} Q(s_{t+1}, a')$, represents the discounted estimate of future rewards. By incorporating both immediate and future rewards, the agent can assess the long-term consequences of its actions, enabling it to pursue strategies that maximize cumulative returns rather than focusing solely on short-term gains. In essence, an action is considered good not only when it yields an immediate reward but also when it leads to more favorable future states. The Q-value for the executed state-action pair $(s_t, a_t)$ is then updated as:

$$Q(s_t, a_t) \leftarrow Q(s_t, a_t) + \alpha \left( \mathrm{TD}_{\mathrm{target}_t} - Q(s_t, a_t) \right). \tag{63}$$

This update increases $Q(s_t, a_t)$ if the target return estimate exceeds the current Q-value, and decreases it otherwise. Over time, $Q(s, a)$ approximates the expected discounted return, guiding the RL agent to prefer parameter configurations that consistently improve the multi-objective search.

Beyond empirical performance gains (as Section 4 shall showcases), our proposed RL-NSGA-II-GRC method provides a theoretical contribution in the form of a principled coupling between dominance-based constrained MOO and RL-based adaptive control. The algorithm can be interpreted as a generational Markov decision process, where the population state is summarized by convergence, feasibility, diversity, and stage indicators, and the RL action defines an adaptive control policy over the evolutionary search dynamics. In particular, the constraint-tolerant dominance introduces a continuum between strict feasibility and controlled constraint relaxation, while the fractional survival sampling introduces a continuum between pure survival and exploratory retention; together, these mechanisms formalize how selection pressure and feasibility enforcement can be adaptively scheduled across generations. Meanwhile, the GRC tie-breaker (in case of same Pareto ranks) provides a weight-free discriminator within a dominance rank, ensuring that Pareto dominance is not overridden while enabling consistent preference toward ideal-point proximity in normalized objective space.

Additionally, from a computational complexity perspective, the proposed RL layer introduces an incremental generational overhead on top of the original NSGA-II, but it does not alter the dominant asymptotic cost terms of NSGA-II. In each generation, the RL agent computes a compact state/reward summary and then performs an ε-greedy action selection and a tabular Q-learning update. The Q-learning decision and/or update step is $O(1)$ per generation (table lookup and a constant number of arithmetic operations). The added cost therefore comes mainly from computing state/reward indicators, that is, the feasibility ratio requires one pass over the population $N$ and is $O(N)$; the diversity indicator is computed over the current non-dominated set and is $O(|\mathcal{F}_0|^2)$; and hypervolume computation depends primarily on the number of objectives $M$ and is typically inexpensive for the not-many-$M$ problems. Separately, our GRC-enhanced tournament only adds scalar scores computation per generation (including the normalization and GRC calculation), which is $O(M)$. The most computationally expensive part is still the NSGA-II backbone, which has the dominant time complexity of $O(MN^2)$ as reported in Abdel-Basset et al. (2018). As a result, comparing the proposed RL-NSGA-II-GRC method against the original NSGA-II, it only increases runtime modestly by a constant factor per generation rather than changing the overall computational order.

**4. Mathematical benchmark MOO problems**

Before applying the proposed RL-NSGA-II-GRC method to a real-world financial application for portfolio optimization, we first evaluate its performance and effectiveness using two well-known mathematical benchmark MOO problems: Kursawe problem (Coello et al., 2007) and CONSTR problem (Deb et al., 2002), which have the true Pareto-optimal fronts reported in literature.

**4.1. Kursawe problem**

Kursawe problem considers a three-dimensional decision vector $\mathbf{x} = (x_1, x_2, x_3)^\top$, with the following two objectives for minimization:

$$\min f_1(\mathbf{x}) = \sum_{i=1}^{n-1} \left( -10 \exp\left( -0.2 \sqrt{x_i^2 + x_{i+1}^2} \right) \right), \quad (64)$$

$$\min f_2(\mathbf{x}) = \sum_{i=1}^{n} \left( |x_i|^a + 5 \sin\left( x_i^b \right) \right), \quad (65)$$

where $a = 0.8$, $b = 3$, and $n = 3$. The decision variables are bounded by:

$$-5 \le x_i \le 5, \qquad i = 1, 2, 3. \quad (66)$$

Coello et al. (2007) stated that there are ways to determine the true Pareto-optimal fronts for certain MOO problems, either theoretically/analytically or with exhaustive computational search (e.g., search by enumerative exhaustive deterministic approach using parallel high-performance computers). Here, the Kursawe's true Pareto-optimal front is openly available at https://emobook.cs.cinvestav.mx/.

We then execute both the original NSGA-II method and the proposed RL-NSGA-II-GRC method (implemented in Python). Both methods are set to run for 300 evolutionary generations and target a final population of 200 solutions. After obtaining the Pareto-optimal fronts from both methods, we compare them with the true Pareto-optimal front. To quantify convergence, we employ the well-known convergence metric (CM), which measures the convergence/closeness of the obtained front to the true front. For each solution, the minimal Euclidean distance to the true front is computed; the CM is then defined as the average of these distances across all $N$ solutions:

$$\mathrm{CM} = \frac{\sum_{i=1}^{N} d_i}{N}, \tag{67}$$

where $d_i$ is the minimal Euclidean distance from each obtained solution $i$ to the true Pareto-optimal front, and $N$ is the total number of obtained solutions (e.g., in this case $N = 200$). A smaller CM indicates a better convergence, as it means the obtained solutions lie closer to the true Pareto-optimal front. In the ideal case, when every obtained solution falls exactly on the true front, CM will be equal to zero.

Figure 2 shows the true Pareto-optimal front of the Kursawe problem alongside the front obtained using the original NSGA-II method. In this case, the CM for NSGA-II is 0.009728. Analogously, Figure 3 displays the true Pareto-optimal front together with the front produced by the proposed RL-NSGA-II-GRC method, which achieves a CM of 0.009169. This corresponds to an improvement in convergence performance of approximately 5.8% compared with the original NSGA-II. To ensure that the observed CM gain is not due to stochastic variability, the problem is repeatedly solved for multiple independent runs with different random seeds while keeping the other setup fixed. We take the mean and standard deviation of CM values across runs for each method and perform statistical significance tests at the 5% level ($p < 0.05$) to evaluate the performance gains, which shows that the improvement of RL-NSGA-II-GRC over the original NSGA-II is statistically significant.

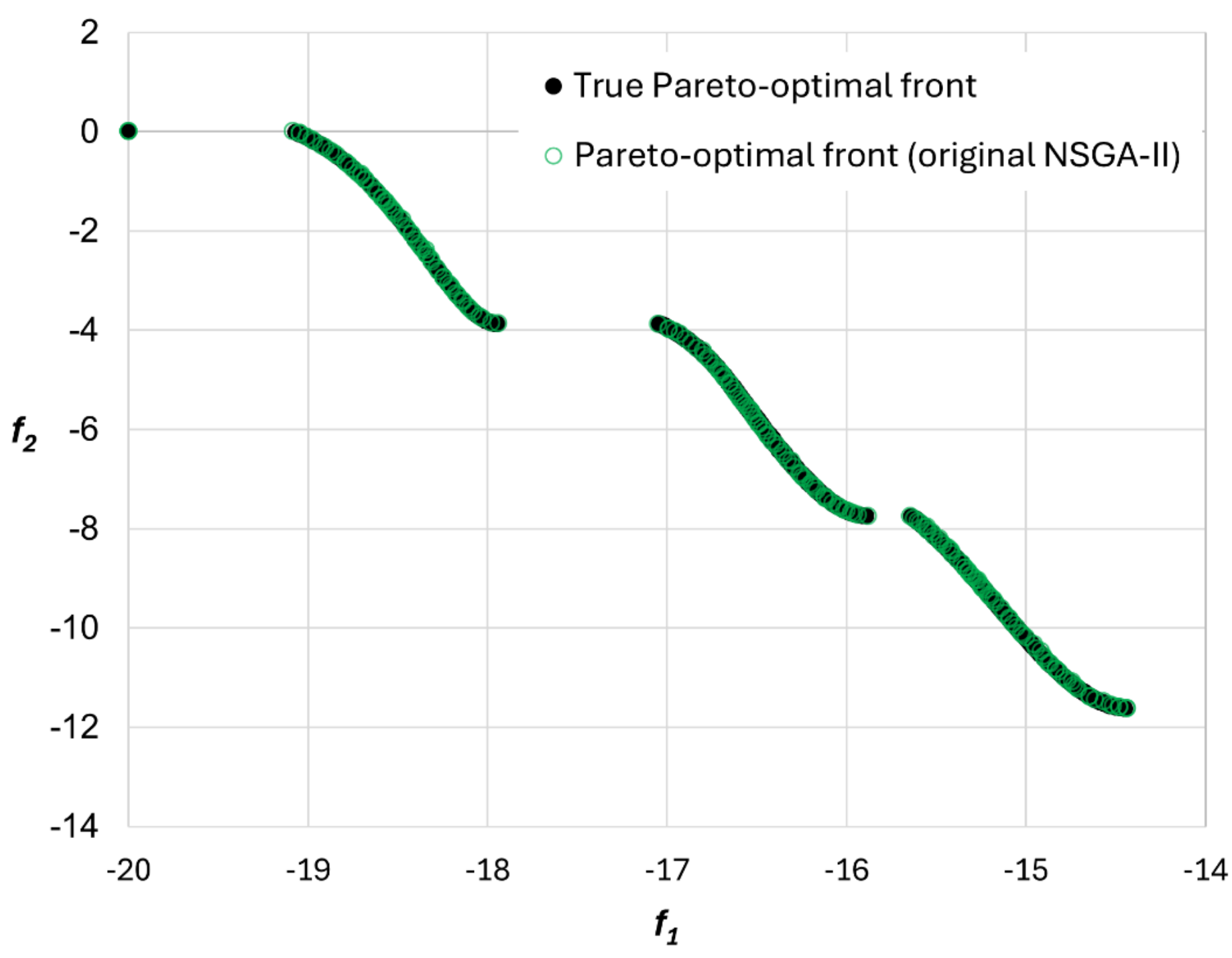

Figure 2. Kursawe MOO problem: the true Pareto-optimal front and the Pareto-optimal front obtained by the original NSGA-II method.

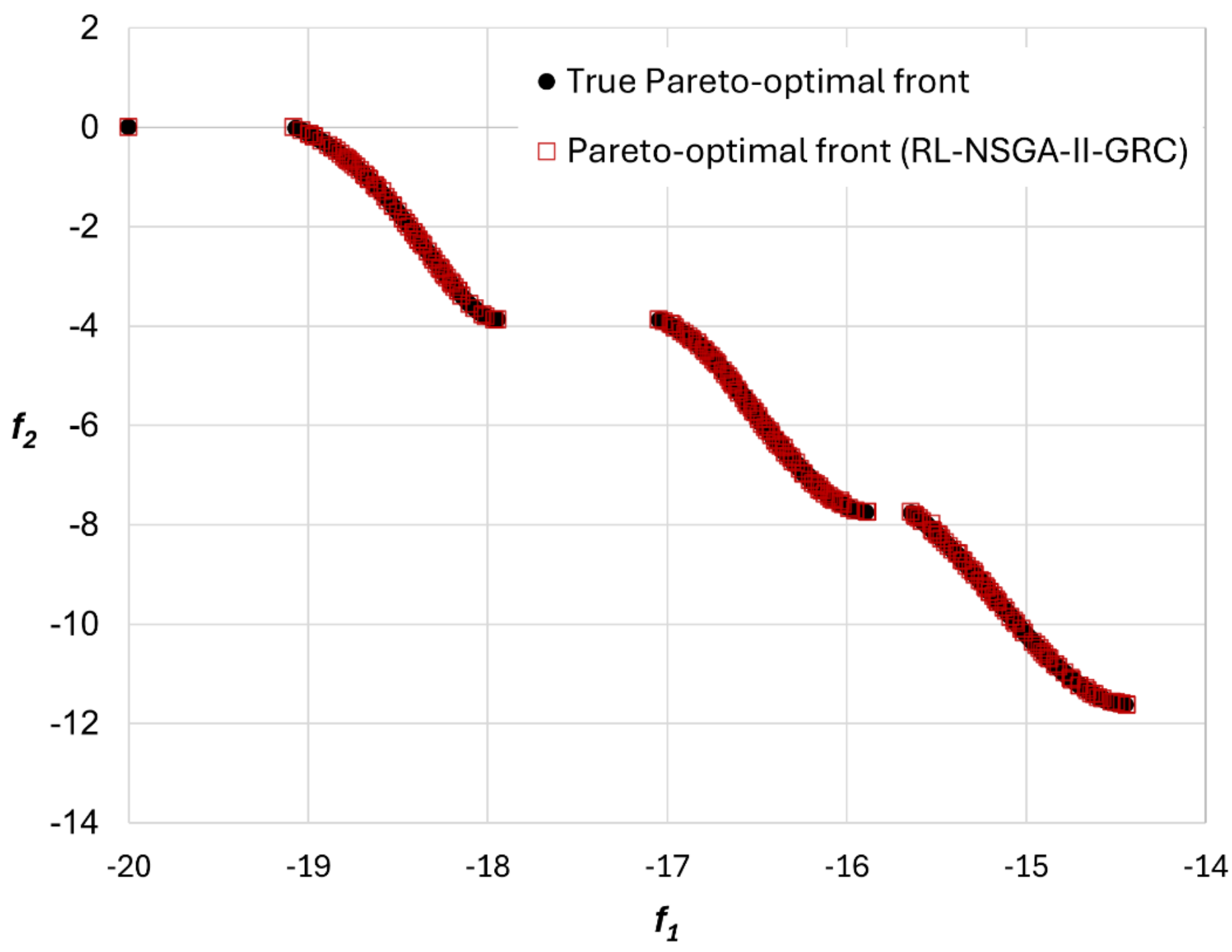

Figure 3. Kursawe MOO problem: the true Pareto-optimal front and the Pareto-optimal front obtained by the proposed RL-NSGA-II-GRC method.

**4.2. CONSTR problem**

The CONSTR problem is another mathematical benchmark MOO problem with a relatively narrow feasible region, making it suitable for testing constraint-handling and exploration-exploitation balance. The problem has two objectives to be minimized:

$$\min f_1(\mathbf{x}) = x_1, \tag{68}$$

$$\min f_2(\mathbf{x}) = \frac{1 + x_2}{x_1}. \tag{69}$$

The variables are subject to the following bounds:

$$0.1 \leq x_1 \leq 1.0, \qquad 0 \leq x_2 \leq 5. \tag{70}$$

In addition, there are two inequality constraints:

$$g_1(\mathbf{x}) = x_2 + 9x_1 - 6 \geq 0, \tag{71}$$

$$g_2(\mathbf{x}) = -x_2 + 9x_1 - 1 \geq 0. \tag{72}$$

The feasible set for the CONSTR problem is therefore given by all $\mathbf{x} = (x_1, x_2)^\top$ satisfying both the bounds and inequality constraints.

Figure 4 illustrates CONSTR's true Pareto-optimal front, together with the fronts obtained by the original NSGA-II and the proposed RL-NSGA-II-GRC methods. Like in Kursawe problem, both methods run 300 evolutionary generations and target a final population of 200 solutions. Results show that the original NSGA-II achieves a CM of 0.01503, whereas the RL-NSGA-II-GRC method achieves a CM of 0.01436. This represents an improvement in convergence performance of approximately 4.4% over the original NSGA-II.

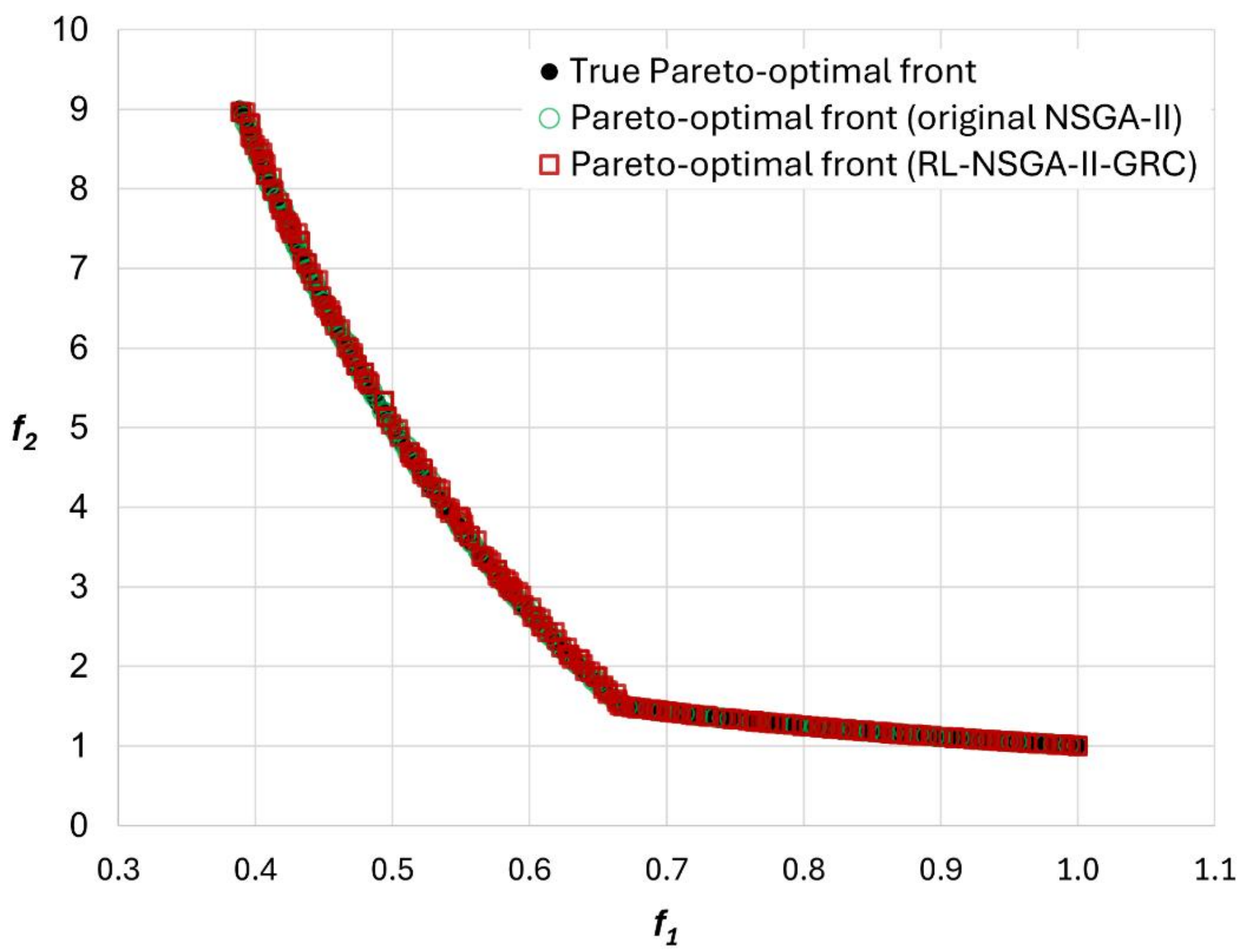

Figure 4. CONSTR MOO problem: the true Pareto-optimal front, as well as the Pareto-optimal fronts obtained by both the original NSGA-II and the proposed RL-NSGA-II-GRC methods.

As observed, the comparative study in this paper focuses primarily on the original NSGA-II because RL-NSGA-II-GRC is designed as a strict enhancement of the NSGA-II backbone. Accordingly, the most direct and controlled evaluation is to compare against the original NSGA-II under the same representation, initialization, constraint set, and computing device, thereby isolating the incremental contributions of the RL-guided adaptive control mechanism, the constraint-tolerant dominance with fractional survival sampling, as well as the GRC-enhanced tournament selection. This one-to-one comparison reduces confounding factors (e.g., differing decomposition strategies or indicator calculations across algorithms) and provides an ablation-like assessment of whether the proposed components improve NSGA-II in a fair and interpretable manner. In addition, NSGA-II remains a widely adopted and competitive baseline in MOO and is frequently used in portfolio optimization studies, making it an appropriate reference point for demonstrating practical improvement.

We also conduct additional ablation experiments under the same settings (e.g., same population size, initialization, and number of generations). Three algorithmic configurations are compared, namely, RL-only (RL-NSGA-II), where the RL agent adaptively adjusts the evolutionary parameters and constraint-handling/survival pressure; GRC-only (NSGA-II-GRC), where NSGA-II is augmented with the GRC-based MCDM ranking in tournament selection; and the full RL-NSGA-II-GRC, which integrates both RL control and GRC-enhanced selection. Across the benchmark problems tested, the ablation results indicate that each component provides a measurable benefit. RL-only variant improves adaptability by dynamically balancing exploration and exploitation, as well as feasibility enforcement across generations, while the GRC-only variant improves selection consistency within the same non-dominated rank by favoring solutions closer to the ideal reference. The full RL-NSGA-II-GRC yields the most consistent overall performance, suggesting a complementary effect between adaptive RL-based control and the GRC-based ranking.

**5. Application: Portfolio optimization with NASDAQ-100 constituents**

To demonstrate the practical usefulness of the proposed RL-NSGA-II-GRC algorithm, we apply it to a real-world portfolio optimization problem. The goal is to design a long-only equity portfolio that simultaneously minimizes risk and maximizes expected return, using recent historical data for large-capitalization stocks from the NASDAQ-100 index.

We start from the 30 largest constituents by weight of the NASDAQ-100 (https://www.slickcharts.com/nasdaq100, accessed in December 2025), and denote their tickers by $\mathcal{A} = \{a_1, a_2, \dots, a_{30}\}$. Daily adjusted closing prices $P_{i,t}$ are downloaded for each asset $a_i$ over the period (from 1 June 2022 to 1 December 2025) using the *yfinance* Python library (https://pypi.org/project/yfinance/, accessed in December 2025). To reduce high frequency noise while retaining enough observations for estimation, the daily prices are resampled to weekly prices by taking the last trading day in each calendar week:

$$P_{i,t}^{(W)} = P_{i,\text{ last trading day of week } t}. \tag{73}$$

From these weekly prices, we compute simple (percentage) weekly returns

$$r_{i,t} = \frac{P_{i,t}^{(W)} - P_{i,t-1}^{(W)}}{P_{i,t-1}^{(W)}}, \qquad i = 1,2, \dots, N, \quad t = 2,3, \dots, T, \tag{74}$$

where $N = 30$ is the number of assets that remain after cleaning and $T$ is the number of valid weeks. Assets with more than 10% missing price data are discarded; remaining gaps are handled by dropping weeks with any missing prices so that all retained series are time-aligned.

Then, using the weekly returns, we compute the mean return $\mu_i$ and the standard deviation $\sigma_i$ for each retained asset $i$, as well as the covariance $\sigma_{ij}$ between the returns of assets $i$ and $j$ (where $i, j = 1,2, \dots, N$). Therefore, we now have a portfolio $p$ consists of $N$ assets with non-negative weights $w_i$ (i.e., fraction of total wealth held in asset $i$). We wish to minimize the risk ($f_1$) and maximize the expected return ($f_2$) of the portfolio simultaneously. Following the Markowitz mean-variance model (Guo, 2022), these two objective functions are thus defined as:

$$\min f_1(\mathbf{w}) = \sum_{i=1}^{N} \sum_{j=1}^{N} w_i \, w_j \sigma_{ij}, \tag{75}$$

$$\max f_2(\mathbf{w}) = \sum_{i=1}^{N} w_i \mu_i \, . \tag{76}$$

The bounds and equality constraint applied to the decision variables (i.e., weights $w_i$) are:

$$\sum_{i=1}^{N} w_i = 1, \qquad 0 \le w_i \le 1, \; i = 1,2, \dots, N. \tag{77}$$

Here, it is worth mentioning that in objective function $f_1$, assets with low or negative covariance ($\sigma_{ij}$) reduce the cross-product terms $w_i w_j \sigma_{ij}$, and thus lower overall risk for a given level of expected return. In other words, the covariance structure explicitly rewards combining assets that do not move together, so the risk objective naturally captures the diversification effect in modern portfolio theory. In addition, as noted earlier, any maximization objective (e.g., the expected return function $f_2$ here) can be readily converted into a minimization objective by multiplying its objective function by $-1$.

For this portfolio optimization problem with 2 objectives and 30 variables, we run the RL-NSGA-II-GRC algorithm for 1000 evolutionary generations, targeting a final set of 1000 Pareto-optimal solutions (although in the end, 989 solutions are found). The resulting Pareto-optimal front (also commonly known as efficient frontier in finance context) is presented in Figure 5. As expected, the efficient frontier clearly exhibits the fundamental trade-off between risk and return, that is, achieving a higher expected return inevitably requires taking on higher risk, while lowering risk leads to a reduction in expected return. This pattern is consistent with finance theory, particularly the Markowitz mean-variance framework, which states that no portfolio can simultaneously minimize risk and maximize return. Instead, we face a set of optimal trade-off choices (i.e., solutions on the efficient frontier). Those located toward the lower-risk end of the frontier are more conservative, offering smaller but more stable returns. On the contrary, those located toward the upper-return end can potentially give us higher returns but also expose us to greater risks.

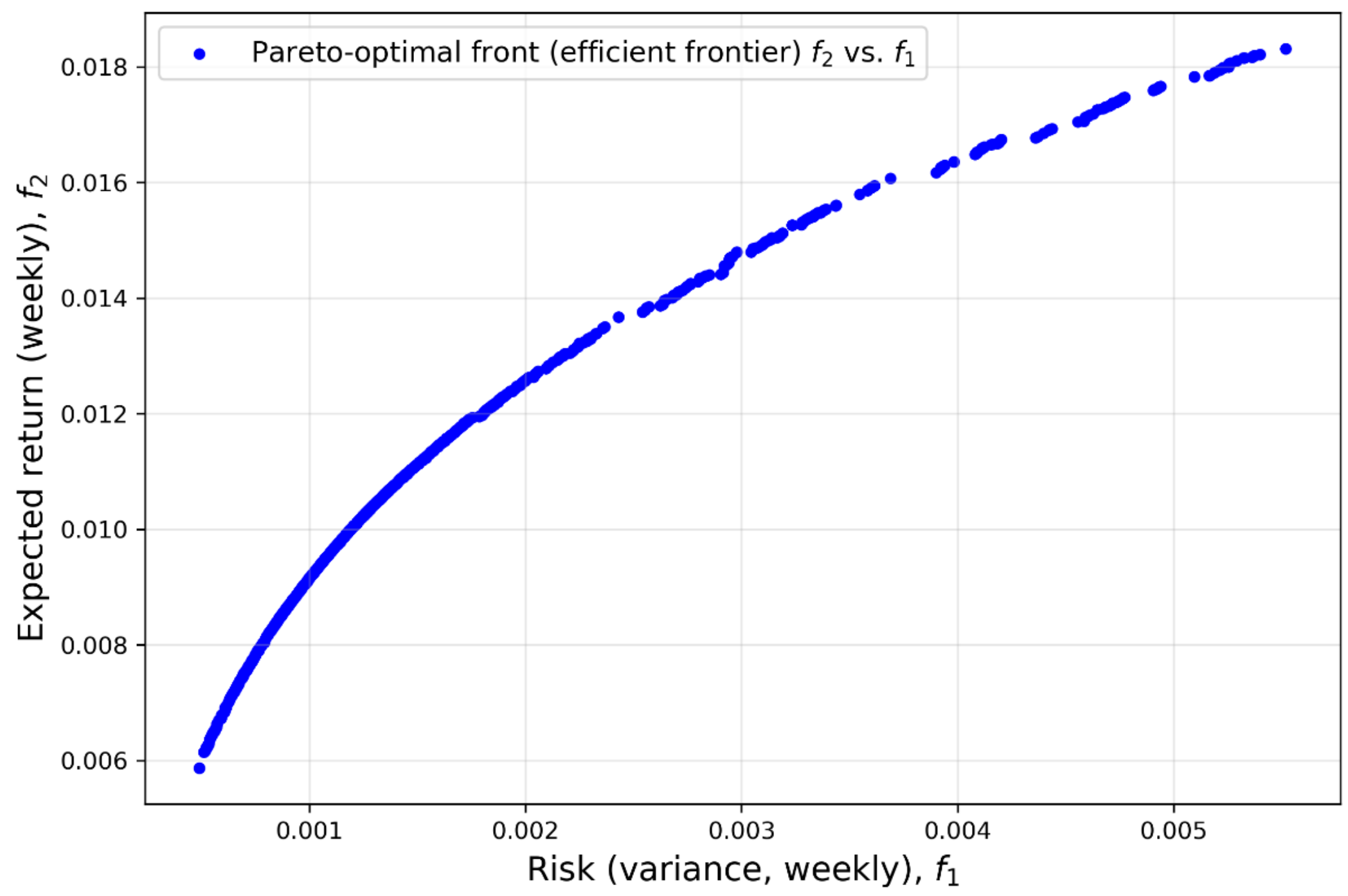

Figure 5. Portfolio optimization problem: efficient frontier ($f_2$ vs. $f_1$) solved using RL-NSGA-II-GRC method.

Additional observations can be drawn from the curvature and smoothness of the frontier in Figure 5. The concave shape indicates diminishing marginal gains in return as risk increases, meaning that there is a progressively larger increment of risk to obtain each additional unit of expected return. Moreover, the dense distribution of Pareto-optimal solutions along the frontier showcases that the RL-NSGA-II-GRC method captures a wide range of efficient portfolio choices. As such, investors with different risk tolerances are able to identify suitable allocations of assets.

Furthermore, with this efficient frontier, we analyse the maximum Sharpe (tangency) portfolio and the utility-maximizing portfolios under different levels of risk aversion. We approximate the weekly risk-free rate $r_f$ using the 3-month U.S. Treasury bill (T-bill) rate, a standard proxy for the USD risk-free asset. As shown in Figure 6, over the period 1 June 2022 - 1 December 2025, the 3-month T-bill yield rose from near 1.0% to more than 5.0% and is currently around 4.0%. The arithmetic average of the 3-month T-bill yield in this period is 4.4578% per annum. The corresponding weekly rate $r_f$ is then calculated as 0.08395% (i.e., 0.0008395).

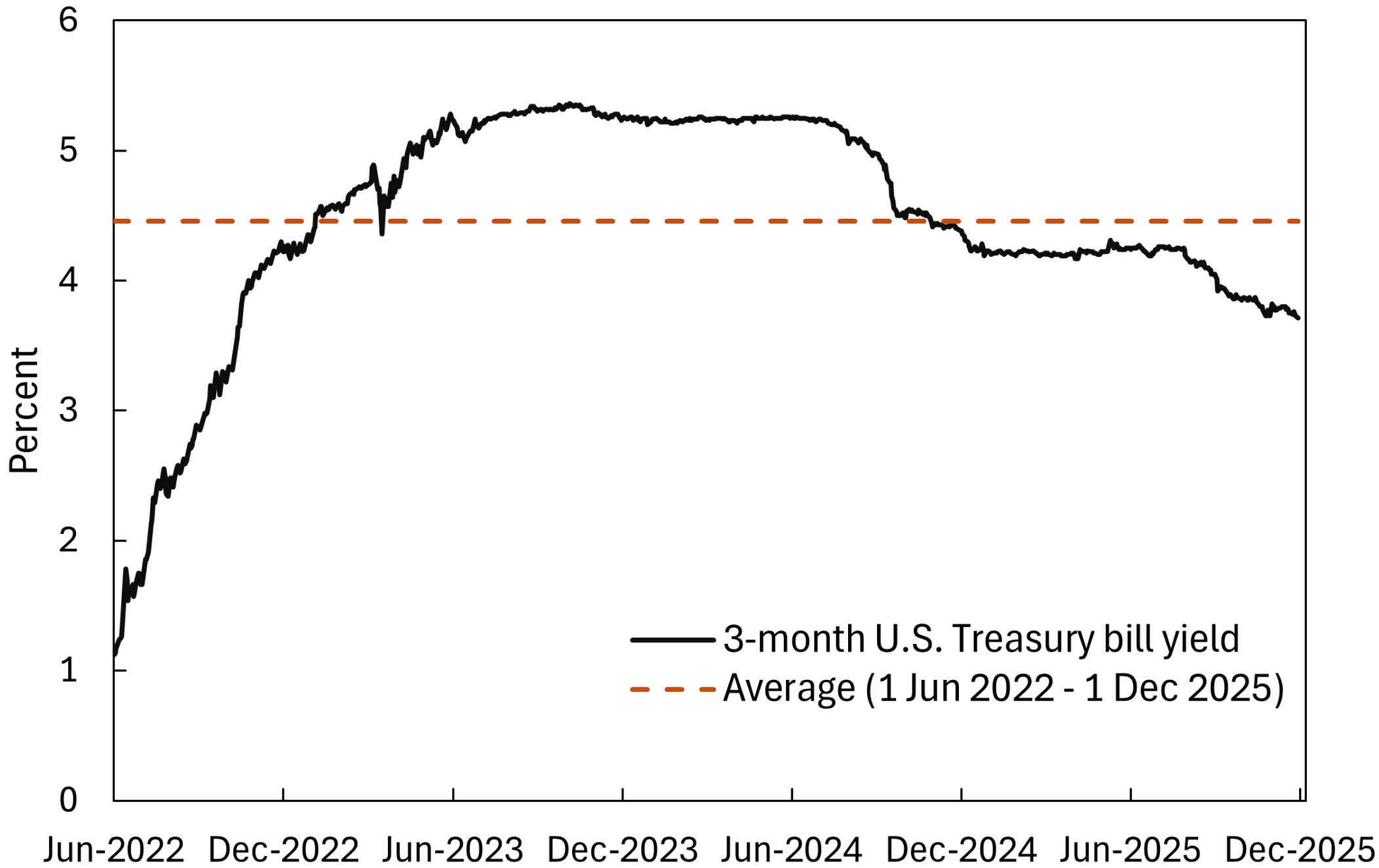

Figure 6. U.S. 3-month Treasury bill yield.

Next, we use $\mu_p$ and $\sigma_p$ to denote, respectively, the expected weekly return ($f_2$) and weekly standard deviation (note: not variance as in $f_1$, but $\sqrt{f_1}$) of a candidate portfolio $p$ on the efficient frontier. For each $p$, we compute the weekly Sharpe ratio

$$S_p = \frac{\mu_p - r_f}{\sigma_p}, \tag{78}$$

and identify the tangency portfolio $p^\star$ as the one with the maximum Sharpe ratio:

$$p^\star = \underset{p}{\operatorname{argmax}} S_p . \tag{79}$$

In this case study, the tangency portfolio has $\sigma_{p^\star} = 0.03765$, $\mu_{p^\star} = 0.01086$, and a weekly Sharpe ratio $S_{p^\star} = 0.2662$ (i.e., annualized Sharpe ratio = 1.92). Financially, this means that among all efficient risky portfolios, $p^\star$ delivers the largest expected excess return (above $r_f$) per unit of risk.

With $p^\star$, the capital market line (CML) can then be constructed:

$$\mu_{\mathrm{CML}}(\sigma) \;=\; r_f + \frac{\mu_{p^\star} - r_f}{\sigma_{p^\star}}\,\sigma, \tag{80}$$

where $\sigma$ is the portfolio standard deviation. The CML, together with the efficient frontier ($f_2$ vs. $\sqrt{f_1}$), are illustrated in Figure 7. As seen, the optimal risky portfolio $p^\star$ corresponds to the tangency point between the efficient frontier and the CML.

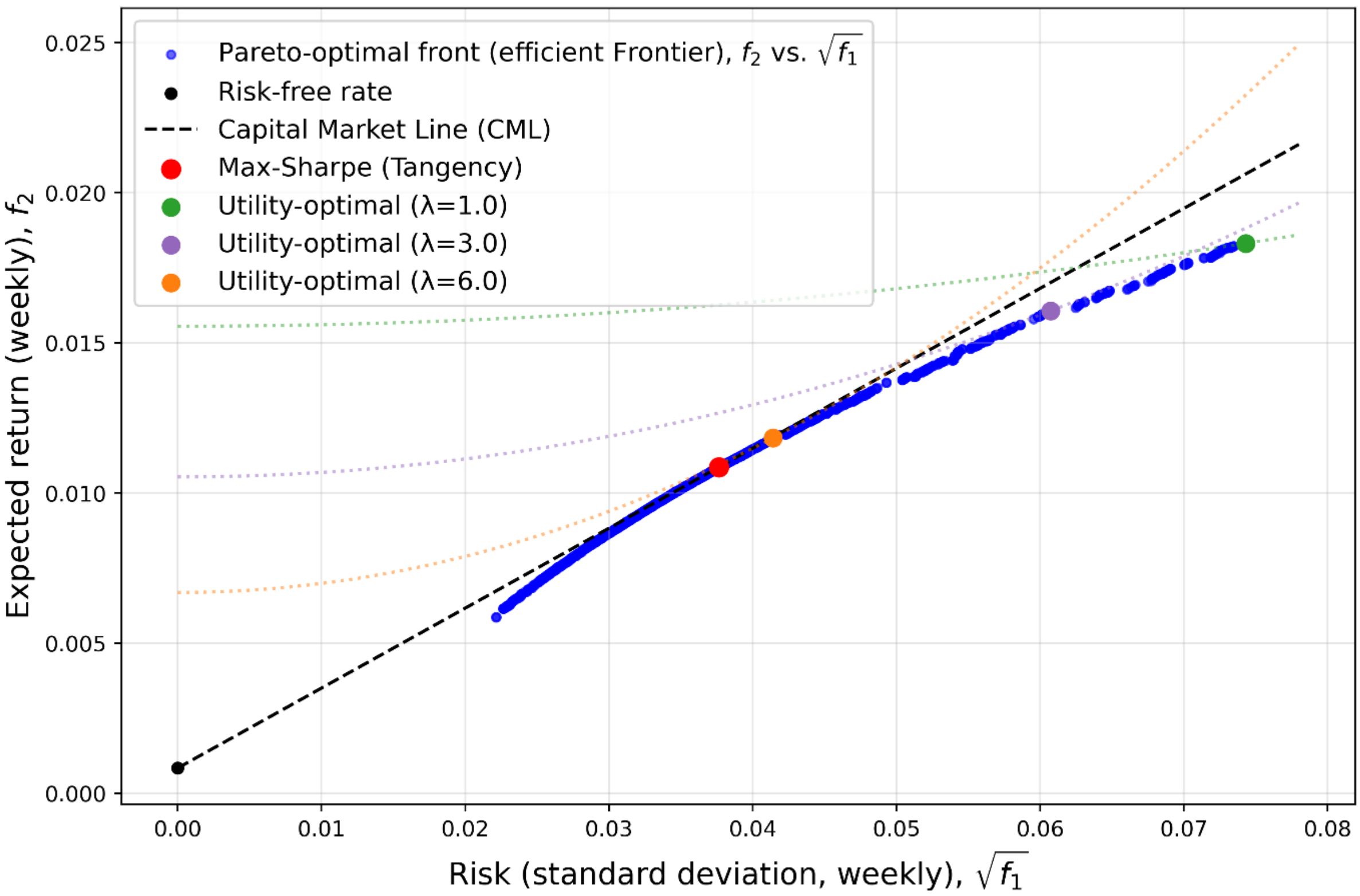

Figure 7. Portfolio optimization problem: efficient frontier ($f_2$ vs. $\sqrt{f_1}$), CML, and risk aversion quadratic utility-optimal points; green, purple, and orange dashed curves are the indifference curves through their respective utility-optimal points.

Moreover, to reflect heterogeneous risk preferences, we also consider a risk aversion quadratic utility function:

$$U(\mu_p, \sigma_p, \lambda) \;=\; \mu_p - \frac{1}{2}\lambda\,\sigma_p^2, \tag{81}$$

where $\lambda > 0$ is the investor's risk-aversion coefficient (larger $\lambda$ indicates stronger aversion to risk). For each prescribed $\lambda$, we evaluate $U(\mu_p, \sigma_p, \lambda)$ for all Pareto-optimal portfolios on the efficient frontier and select the utility-optimal portfolio:

$$p^{\star}(\lambda) = \underset{p}{\operatorname{argmax}}\, U(\mu_p, \sigma_p, \lambda). \tag{82}$$

For instance, as demonstrated in Figure 7, we examine $\lambda \in \{1,3,6\}$, corresponding to low, moderate, and high risk aversion, respectively. As expected, the resulting utility-optimal portfolios exhibit decreasing risk and expected return as $\lambda$ increases.Specifically, for $\lambda = 1$ we obtain $\sigma_p = 0.07429$, $\mu_p = 0.01831$; for $\lambda = 3$, $\sigma_p = 0.06074$, $\mu_p = 0.01607$; and for $\lambda = 6$, $\sigma_p = 0.04140$, $\mu_p = 0.01183$; these portfolios are shown as the green, purple, and orange points, respectively, in Figure 7.

**6. Discussion and Limitations**

The numerical experiments on the Kursawe and CONSTR benchmarks, together with the NASDAQ-100 case study, show that our proposed RL-NSGA-II-GRC method is both effective and versatile for MOO problems. On the two benchmark problems with known true Pareto fronts, RL-NSGA-II-GRC consistently yields Pareto-optimal fronts that lie closer to the true front than those produced by the original NSGA-II, as reflected in the reduction of the CM by about 5.8% for Kursawe and 4.4% for CONSTR. At the same time, the constraint-tolerant dominance scheme and fractional sampling survival selection preserve a well-spread and diverse set of non-dominated solutions, instead of collapsing to a narrow region of the front. The RL agent plays a key role here: by adaptively tuning crossover and mutation probabilities, the constraint tolerance, and the front sampling fraction in response to hypervolume, feasibility, and diversity feedback, it automatically balances exploration and exploitation throughout the generations. This adaptive capability is significantly bolstered by the introduction of the GRC, which ensures that selection pressure simultaneously accounts for Pareto rank, crowding distance, and geometric proximity to ideal reference values.

Consequently, the resulting Pareto-optimal fronts are not only closer to the theoretical optima but also smoother and more uniformly distributed, highlighting the synergy between dynamic RL agent control and GRC-based parent tournament selection.

Further, to examine the sensitivity of the RL-NSGA-II-GRC framework to varied design choices, we additionally evaluated alternative discretization granularities and alternative reward-weight configurations. Specifically, we considered (i) finer and coarser binning for feasibility and diversity levels, and (ii) multiple reward weight triplets that varied the relative emphasis placed on hypervolume versus feasibility/diversity while keeping all other algorithmic components fixed. The overall trends observed in the benchmarks and the portfolio case indicate that the algorithm's performance is sensitive (but not overly) to moderate variations of these settings: the controller continues to learn consistent action preferences (e.g., stronger exploration early and stricter feasibility enforcement later), and the resulting Pareto-front quality remains comparable across a reasonable range of discretization and weight choices. Nevertheless, extremely coarse discretization may reduce the controller's ability to react to intermediate search conditions, whereas overly fine discretization may dilute learning by expanding the state space. Similarly, setting feasibility/diversity weights too close to zero can weaken feasibility enforcement or reduce spread. These findings suggest that the proposed default settings provide a balanced trade-off, while the design definitely remains tunable for problem-specific requirements.

For the time-complexity analysis of the proposed RL-NSGA-II-GRC against the original NSGA-II, as the population size $N$ increases, the dominant runtime within RL-NSGA-II-GRC is consistently driven by non-dominated sorting of NSGA-II, i.e., on the order of $O(MN^2)$. The RL component adds $O(N)$ feasibility bookkeeping and up to $O(|\mathcal{F}_0|^2)$ diversity computation per generation, plus an $O(1)$ tabular update. The GRC component only adds an inexpensive scalar score computation and does not change the dominant complexity term. Thus, the RL and GRC overhead remain secondary, and the overall computational order continues to be governed by the NSGA-II backbone. This is also further corroborated by our empirical observations in both the benchmark mathematical problems and the NASDAQ portfolio optimization problem, where execution times for RL-NSGA-II-GRC and the original NSGA-II do not significantly deviate. Ultimately, the RL and GRC layers contribute only a modest constant factor overhead rather than shifting the computational complexity of the algorithm.

Regarding performance relative to other state-of-the-art MOEAs (e.g., decomposition-based or indicator-based methods, as well as many-objective extensions), the present study does not claim that RL-NSGA-II-GRC universally dominates all alternatives; rather, it demonstrates that the proposed RL-controlled mechanisms and the GRC tie-breaking strategy can significantly strengthen the widely used NSGA-II backbone for MOO problems. Importantly, the proposed framework is largely modular, which means that the RL agent (our state-action-reward design) and the GRC-based ranking can, in principle, be integrated with other evolutionary algorithms by treating their control components (e.g., variation strength, constraint relaxation, or selection pressure) as RL actions and using convergence, feasibility, and diversity indicators as feedback. For hybrid RL-based optimizers, RL-NSGA-II-GRC differs in that it preserves a Pareto-first ranking logic and uses GRC when dominance information is insufficient, while also explicitly controlling constraint tolerance and survival pressure across generations. A comprehensive head-to-head comparison with additional state-of-the-art MOEAs and recent RL-assisted optimization algorithms is therefore a valuable direction for future work and will be pursued in subsequent studies.

The real-world NASDAQ-100 portfolio application further underlines the practical value of the proposed RL-NSGA-II-GRC. The algorithm constructs a smooth, concave efficient frontier in the mean-variance space, with well-populated trade-off regions that make it easy to identify the tangency (i.e., maximum Sharpe ratio) portfolio and utility-optimal portfolios for different levels of risk aversion. This indicates that the RL guided NSGA-II with GRC are able to navigate a realistically constrained financial search space and uncover portfolios that are both attractive and interpretable for decision makers. However, this work is not without limitations. One limitation lies in the design of the RL agent. Our current implementation adopts a tabular Q-learning agent with a discrete state-action space, which is transparent and effective for all the problems studied here, but could be further extended through function approximation (e.g., deep RL) and more expressive state features to handle even higher dimensional, noisier, or more dynamically changing MOO environments. This would be a promising direction for future work.

## 7. Conclusion

In conclusion, this study proposed a RL-guided NSGA-II method enhanced GRC (RL-NSGA-II-GRC) method to address constrained MOO problems and demonstrated its application to NASDAQ portfolio optimization. The method integrated a tabular Q-learning agent into the NSGA-II backbone to adaptively control key evolutionary parameters, including crossover probability, mutation strength, constraint tolerance, and front sampling fraction, based on population-level feedback from hypervolume, feasibility ratio, and diversity indicators. In parallel, a GRC-enhanced binary tournament selection operator was designed to assess potential parents using a unified performance score that simultaneously accounted for dominance rank, crowding distance, and geometric proximity to ideal objective values, thereby strengthening selection pressure toward high-quality, well-distributed solutions.

The numerical experiments on the Kursawe and CONSTR benchmark MOO problems showed that RL-NSGA-II-GRC achieved better convergence to the true Pareto-optimal fronts than the original NSGA-II, with CM improvements of approximately 5.8% and 4.4%, respectively, while preserving a diverse and well-spread set of non-dominated solutions. In the NASDAQ portfolio case study, the proposed method produced a smooth, concave efficient frontier in the mean-variance space with a dense distribution of Pareto-optimal portfolios, enabling clear identification of the portfolio with maximum Sharpe ratio and utility-optimal portfolios for different levels of risk aversion. These results indicated that the combination of RL agent, constraint-tolerant dominance, fractional front sampling, and GRC-based selection effectively balanced exploration and exploitation and was capable of navigating realistically constrained financial search spaces to yield practically meaningful solutions. Although the proposed framework performed well across both benchmark and real-world problems, there remained room for further enhancement. In particular, the tabular Q-learning agent with a discrete state-action space, while transparent and effective for the settings studied, could be extended to more expressive function-approximation schemes and richer state representations to better handle higher-dimensional, noisier, or more dynamically evolving optimization environments. Future research may also investigate alternative reward designs, additional performance indicators, and broader classes of risk measures and portfolio constraints, as well as applications of RL-NSGA-II-GRC to other optimization and decision-making domains where constrained multi-objective trade-offs are of central importance.

**Acknowledgment**

We thank Prof. Carlos A. Coello Coello for providing the valuable true Pareto-optimal fronts for MOO benchmark problems.